# Modelling of Received Signals in Molecular Communication Systems based machine learning: Comparison of azure machine learning and Python tools

Soha Mohamed[1], Mahmoud S. Fayed[2]

[1] School of Computer Science and Technology, Harbin Institute of Technology, Harbin 150001, China.
[2] College of Computer and Information Sciences, King Saud University, Riyadh, Saudi Arabia.

**Abstract**

Molecular communication (MC) implemented on Nano networks has extremely attractive characteristics in terms of energy efficiency, dependability, and robustness. Even though, the impact of incredibly slow molecule diffusion and high variability environments remains unknown. Analysis and designs of communication systems usually rely on developing mathematical models that describe the communication channel. However, the underlying channel models are unknown in some systems, such as MC systems, where chemical signals are used to transfer information. In these cases, a new method to analyze and design is needed. In this paper, we concentrate on one critical aspect of the MC system, modelling MC received signal until time $t$, and demonstrate that using tools from ML makes it promising to train detectors that can be executed well without any information about the channel model. Machine learning (ML) is one of the intelligent methodologies that has shown promising results in the domain. This paper applies Azure Machine Learning (Azure ML) for flexible pavement maintenance regressions problems and solutions. For prediction, four parameters are used as inputs: the receiver radius, transmitter radius, distance between receiver and transmitter, and diffusion coefficient, while the output is mAP (mean average precision) of the received signal. Azure ML enables algorithms that can learn from data and experiences and accomplish tasks without having to be coded. In the established Azure ML, the regression algorithms such as, boost decision tree regression, Bayesian linear regression, neural network, and decision forest regression are selected. The best performance is chosen as an optimality criterion. Finally, a comparison that shows the potential benefits of Azure ML tool over programmed based tool (Python), used by developers on local PCs, is demonstrated.

1. INTRODUCTION

Nano-technology facilitate the development of nanometer-scale devices [1]. A Nano machine is the most fundamental primary component for performing basic occupations such as computing, data storage, actuation, and sensing. Communication between Nano-machines will increase the complexity and range of operation of individual nano-machines' capabilities and applications [2].

Molecular communication (MC) is the most promising technique for nanoscale communication. Molecules in MC transmit information across an aqueous medium [3].

The information can be encoded in terms of time of release, concentration, or molecule type [4]. Nano network, a future framework, is designed with the inter-connection of Nano-machines. Modeling of Nano networks has been proposed in biomedical, environmental, and military applications [5].

The Internet of Bio-Nano Things (IoBNT) is presented as a network of biological components and nano-machines that communicate with one another and the internet to empower different applications such as healthcare-monitoring. Additionally, MC has emerged as the most capable nano-scale communication prototypical for IoBNT [6].

Moreover, it has been identified as a potential upcoming application in the health field, and we can see this in disease monitoring/therapy/diagnosis [7] as well as target drug-delivery (TDD) [8].

Detection techniques are very important for the trustworthy retrieval of the signal directed from a communication source, in which the concentration encoded signal is calculated from the degraded variety detected at the decoder [9]. Detection technique design and analysis have traditionally relied on computational models that describe the transmission process, signal and receiver noise propagation, and many other components of end-to-end signal transmission and reception [10].

MC relies on chemical signal and doesn't use electromagnetic (EM) signals, and the consistent source model may be unidentified or mathematically irreconcilable. One solution to this problem is using detection techniques inspired by ML [11] which can be used to design detection techniques that can learn directly from data.

It is clear that there are no research creativities in the field of MC-based signal detection of the received signal until time $t$ using automated ML. In a production-ready environment, model automation is frequently disregarded. As a result, we investigate automation in the proposed work to make the model accessible to the end-user. Whenever there is a requirement to run computationally complex ML models, automation is usually required. For automation, one viable solution to examine is ML as a service (MLaaS) [12]. Automated ML implies that scientists do not need to create a logistic regression or a neural network model from zero but may utilize the benefits of automated algorithmic components offered on MLaaS [13]. As a proof of concept, the work of automation is accomplished utilizing the MLaaS model. Typically, the data which ML programs rely on it is challenging to process and store. It affords a robust and effective crossing point that allows ML specialists to carry out experiments with their datasets rather than worrying about storage, compute, or networking issues [14]. When the difficulty of data grows, ML poses numerous difficulties. As a result, automating an ML activity becomes critical.

Hyper-Parameter Optimization (HPO) is without a doubt one of the most important aspects of automated ML [15], which is regularly challenging to attain. As a result, in the proposed scheme, we used Azure ML, a production-ready tool, to ease designing ML infrastructure. Essentially, it is critical to recognize a significant variance between implementing algorithms on a device and software as a service portal. However, it is worth noting that all of these algorithms are performed on an MLaaS prototype, which allows for extensive fully automated tuning of hyper-parameters, significantly lowering human work and time. Azure ML, when combined with ML Operations, aids in the improvement of workflow performance. Traditional ML frameworks' credibility in a real-time scenario involving massive networking data is highly dubious. However, if the focus is given on the use of MLaaS settings such as BigML, Algorithmic, Data

Robot [16], and Azure ML (as in the proposed work), it can greatly assist in achieving load balancing or auto-scaling [17], [18].

In this study, MC received signal is modeled with different regression algorithms in Azure ML. The following regression algorithms are applied: Bayesian regression, neural network regression, random forest regression, and boosted decision tree regression. According to the experimental results, the best method is identified. In this work, it is illustrated that establishing an end-to-end MC system on Azure ML is a simple and efficient task with numerous algorithms to choose from. Also, a comparison between Azure ML as a visual tool and Python as textual- codes based tool is shown.

## 2. Overview and Related works of the Visual programming language (VPL)

Software development is critical in today's age of information technology for meeting the needs of businesses and organizations for high-quality information systems. Creating high-quality, dependable, scalable, efficient, user-friendly, and affordable software is not easy. Thus, there is an increasing request for software development tools that are simple to learn, versatile, and productive [19]. Various components of the software development process have changed due to the sophistication of software requirements, and many tools have been invented to assist programmers. Large projects require integrated development environments (IDEs) such as NetBeans, Eclipse, and Microsoft Visual Studio. While using these tools, programmers must understand the general programming paradigms and the rigorous syntactic of each scripting language used. The program representation in the source code of these tools is limited to text, as photos and graphics cannot be included. Furthermore, the more sensitive the programming language is, the more complicated its syntax becomes, making it harder to understand and code. This contest made it more attractive to use VPLs, which catch more programmers and increase software development productivity [20].

VPLs allow the creation of computer programs and applications by means of multi-dimensional visual representation, and they give a programming system based on communication with graphical features that combine Text, Shapes, Colors, and Time, rather than writing textual code-base [21]. VPLs enhance the techniques through which programmer's express information presentation and processing. They employ images, animations, drawings, and icons to help users understand and change the outcomes visually [22].

There are a lot of VPLs on the market right now, but the most popular and commonly used ones are academic programs like Scratch, Alice, and Kodu, or domain-specific tools like Azure ML, and LabVIEW (Data Acquisition, Instrument Control, and Industrial Automation). Wide-ranging-purpose VPL such as Limnor, Tersus, and Envision [22].

We will go over a few of the most successful VPLs in the next subsection. Forms/3 is an example of a form-based VPL; diagrammatic VPLs include LabVIEW and Tersus, and icon-based VPLs include Limnor and Kodu [23].

Scratch is a program developed by the MIT Media Lab. It is a novel VPL platform that allows you to make interactive stories, games, cartoons, songs, and artwork.

Scratch allows instructors to minimize students' mental workload when they first begin learning to program [24].

Kodu is a teaching tool. Microsoft Research created. It is a brand-innovative icon-based VPL designed exclusively for video game production. It covers the fundamentals of game production. It is intended to be user-friendly for kids and fun for all. Kodu is event-driven, with each line of code consisting of a circumstance and an activity [25].

Alice is a VPL that allows users to modify objects in 3D environments. Carnegie Mellon University was responsible for its creation. It allows students to know about programming ideas based object-oriented without dealing with the grammar issues that text-based programming languages might cause. A program writer in a GUI setting uses Alice to choose programed constructions and method commands from a list of options [26].

Scratch, Kodu, and Alice are only used for educational purposes. Furthermore, at the student level, Scratch and Kodu are limited. On the other hand, Alice can be utilized by teachers and other inexperienced programmers to investigate computational reasoning principles [27].

LabVIEW is a diagrammatic VPL designed for scientists and engineering; it is not a free program. It is a visual platform with built-in data capture, process control, measurements analysis, and descriptive statistical features [28]. The operation of LabVIEW is decided by a layout of the graphical block diagram on which the program writer links diverse functions node by wires [29].

Tersus is another form of a diagrammatic VPL. It was established by Tersus Software Ltd.'s. It is designed to help developers create sophisticated web-based applications by graphically describing operator interfaces, user behavior, and server operations. It's a general-purpose programming language for creating flow diagrams and interacting with applications. [30].

Limnor is a VPL for Visual Studio.NET that is based on icons. It developed by Long Flow Enterprises Limited. It can make various applications, including online interactive kiosks, sales promotions, database management systems with responsive query, and business administration systems. This tool can be used by non-technical individuals [31]. Limnor, unlike the other tools, is not freeware and requires payment to obtain a copy. Furthermore, Limnor exclusively generates code in C#.

PWCT (programming without coding) is general purpose visual language that support textual code generation in many programming languages including Python (So can be used in ML applications based on Python libraries and frameworks) [22].

3. **System model**

The system model, which we proposed in [32], is used in this study. As shown in Figure 1, both the transmitter and the receiver have unique radiuses and are spherical surfaces. This study is focused on a trapezoidal-container. As a result, molecules in the container will float from the transmitter to the receiver, touching the receiver's surface to the transmitter's surface. The sphere's sample points construct the molecules at random on the transmitter sphere dimension. The molecules are supposed to represent a random walk through the container. with the positions changing at random because noise is introduced to the molecule's location as presented in equation 1 and 2.

$$\begin{cases} x \to x + \Delta x \\ y \to y + \Delta y \\ z \to y + \Delta z \end{cases} \tag{1}$$

$$\begin{cases} \Delta x \sim N\ (0.2D\Delta) \\ \Delta y \sim N\ (0.2D\Delta) \\ \Delta z \sim N\ (0.2D\Delta) \end{cases} \tag{2}$$

where D is the coefficient of Diffusion, and $\Delta \to$ discrete time step.

A Gaussian distribution is used to generate the random noise, with mean = 0, SD (standard deviation) = 2 * D *.

The striking molecules are reflected back into the container at the container surface.

$$\begin{cases} (x_1.y_1.z_1) \leftarrow Previous\ Position \\ (x_2.y_2.z_2) \leftarrow Position\ after\ striking\ container \\ (x_f.y_f.z_f) \leftarrow Reflected\ Position \end{cases} \tag{3}$$

$$\begin{cases} a = (x_2 - x_1)^2 + (y_2 - y_1)^2 \\ b = 2\ (x_2 - x_1)x_1 + (y_2 - y_1)y_1) \\ c = x_1^2 + x_1^2 - R_v^2 \end{cases} \tag{4}$$

where $R_v^2$ is the radius of the vessel.

$$at^2 + bt + c = 0 \to t_1\ and\ t_2 \tag{5}$$

In this circumstance, $t_1$ and $t_2$ are the quadratic equation origins

$$\begin{aligned} x_i^1 &= x_1 + (x_2 - x_1) * t_1 \\ y_i^1 &= y_1 + (y_2 - y_1) * t_1 \\ x_i^2 &= x_1 + (x_2 - x_1) * t_2 \\ y_i^2 &= y_1 + (y_2 - y_1) * t_2 \end{aligned} \tag{6}$$

$$Distance\ 1 = (x_i^1 - x_1)^2 += (y_i^1 - y_1)^2$$
$$Distance\ 2 = (x_i^2 - x_1)^2 += (y_i^1 - y_1)^2$$

Due to the fact that it is the point inside the container, the solution with the shortest distance is picked.

$$\begin{aligned} x_f &= 2 * x - x_2 \\ y_f &= 2y - y_2 \\ z_f &= z_2 \end{aligned} \tag{7}$$

The reflection is only in the $x$ and $y$ directions, $z$ will remain unaffected according to the equations above. As the molecules float, their positions change, causing some to collide with the container's surface. Likewise, there is an option to determine a variable numerically, such as with the particle's reflecting location after colliding with the container. The procedure will continue until the molecules reach the receiver's surface. The molecules are absorbed when they come into contact with the receiver surface. As a result, the container and receiver may absorb the same number of molecules cumulatively. In this scenario, the procedure can go on for up to 3000 rounds. The probability is that an accumulated precision of molecules striking the receiver's surface is determined, which is performed by adding the number of molecules striking the surface at all-time $t$, and then dividing the value by the entire number of molecules. After 3000 rounds, this technique returns the percentage or mean average precision (mAP). After that, the mAP is then derived by taking the average of the precisions.

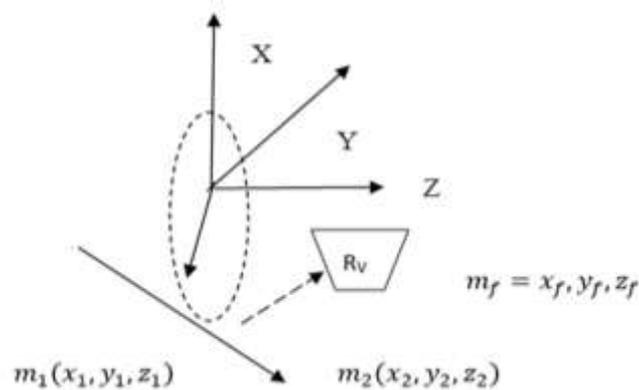

Fig.1 The system model.

## 4. METHODOLOGY

### A. Objective

The experiment's goal is to create an Azure ML tool and compare different kinds of regression ML models for signal detection in MC, then we choose models with higher performance for further investigation. Moreover, we compare the Azure ML (as a visual language) to code-based language like Python.

### B. Experimentation Platform

Microsoft Azure ML Studio "(https://studio.azureml.net)" was chosen to develop a tool and run experimental studies. The Azure MLS is chosen because of the following characteristics that contribute directly to the study's objectives:

- ML as a service delivered via the cloud.
- Browser-based solution – the user only requires a browser to interact with the system. There are no issues with set-up, installation, or repairs.
- Simple drag-and-drop interface for combining computer elements into an investigation in a logical and straightforward manner.

- There are many designed regression modules that are available to be using.
- The ability to code experiments with the Python and R languages.
- R package functions can be integrated.
- Integrated development environment that is password-protected and easy to use.
- Ability to distribute results of the experimentations to the web.
- The ability to reuse previously published experiments or parts of them.
- Low or no cost (pay as you go) service.

Azure Machine Learning [33] is a cloud-based solution that allows the implementation of ML procedures. Azure is a public cloud platform from Microsoft. The following are some of the advantages of adopting the public cloud computing platform (Azure ML): processing large amounts of data and allowing the admission from any location. The procedure of Azure ML is shown in Figure 2. Azure ML comes with a graphical tool for managing the ML process, data pre-processing modules, a set of ML algorithms, and an API for launching models into apps. Azure ML is a graphic tool that allows the user to regulate the procedure from start to finish, including data pre-processing, running experiments with ML approaches, and testing the final model. MLS also assists customers in deploying their models to the clouds.

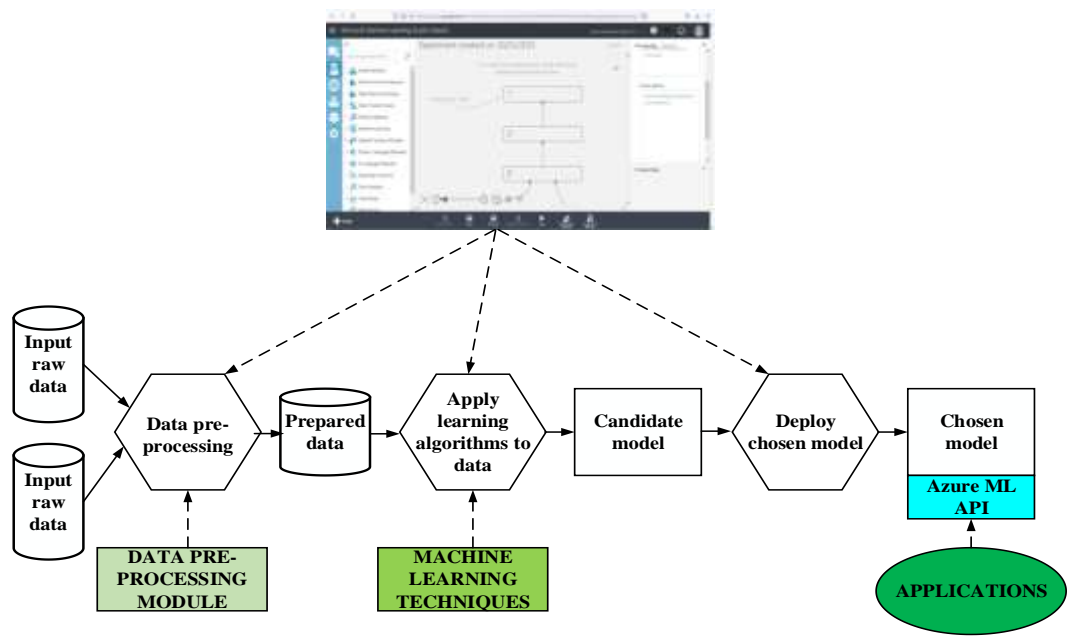

Fig.2 Block diagram of the azure ML experiment procedure.

The following are the Azure ML components:

- A graphical tool that can be used to manage the entire process from start to finish. The ML team can use this tool to apply pre-processing data modules to raw data, perform experiments using an ML algorithm on the prepared data, and test the resulting model. When a good model is found, MLS assists its users in deploying it on Microsoft Azure.
- A collection of modules for data preparation.
- A collection of ML algorithms.
- An Azure ML API that allows apps to access the specified model after being installed on Azure.

## C. Generation of training and testing data

There is only one mAP per simulation, and the simulation is specified by the simulation parameters. The radius of the receiver ($r_R$), the radius of the transmitter ($r_T$), the distance between the receiver and the transmitter ($d$), and the diffusion coefficient ($D$) are among these characteristics. Table 1, the selected model parameters are utilized to forecast mAP (mean average precision).

Table 1 Model parameters to predict mAP.

| Training Data | | Prediction |
|---|---|---|
| [$R_T$, $R_R$, d, D] | → | mAP |

This simulation is essentially run with various simulation parameter, resulting in various mAP. The models are then developed, which map the simulation parameters with the determined mean average precision. The data is acquired by altering the various system settings and running the simulation to obtain each value of mAP, as shown in Figure 3. After obtaining the first set of findings, data was collected by modifying the system's parameters and running the simulation again to obtain the mAP values for each dataset.

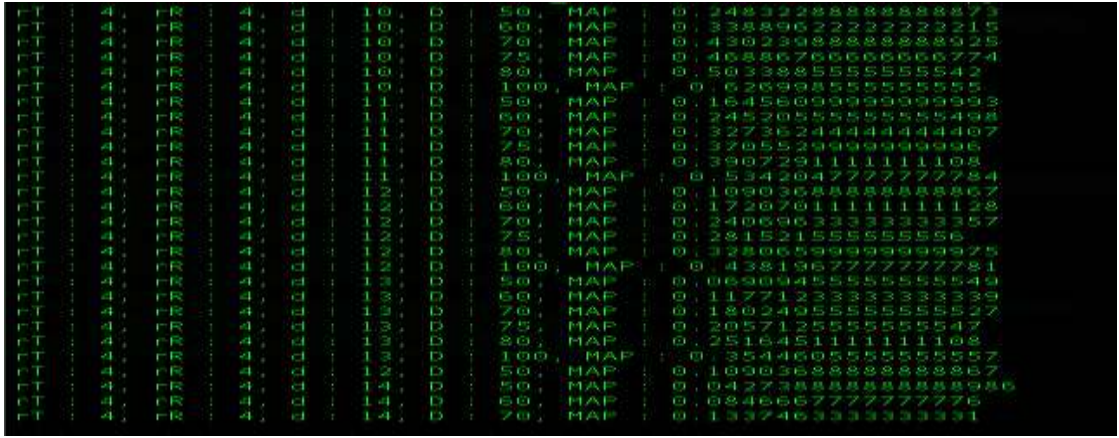

Fig.3 A figure depicting the creation of mAP values.

In Table 2, we present the system parameters and their values that we used for the simulation.

Table 2 the range of parameters used for the simulation.

| Parameter | Variable | Range of value |
|---|---|---|
| Diffusion coefficient | D | {50,60,70,75,80,100} |
| The radius of the receiver | $R_R$ | {4,5,6,7.5,8,10} |
| The radius of the transmitter | $T_R$ | {4,5,6,7.5,8,10} |
| Number of molecules | N | 1000 |
| Distance | d | {2,3,4,5,6,7,8,9,10,11} |
| Time step | $\Delta t$ | 0.01 |

D. **Regression models included in the experiment**

Models that come pre-installed with Azure:-

   *a) Bayesian Linear Regression*

In the Bayesian technique, Bayesian inference is applied [34]. Equations 8-9-10-11 are used to determine the normal distribution in the Bayesian technique. Due to the fact that, $w$ is random variable with a continuous value in $R^d$, According to the Bayes rule, $y$ is the posterior distribution of $w$.

$$P(w|y) \propto P(y|w) P(w) \qquad (8)$$

$$P(w|y) \sim N(\mu, S) \qquad (9)$$

$$S^{-1} = S_0^{-1} + \frac{1}{\sigma^2}(x^t) \qquad (10)$$

$$\mu = S(S_0^{-1} M_0 + \frac{1}{\sigma^2} x^t y) \qquad (11)$$

The predictive distribution in Bayesian linear regression is computed using Equation 12:

$$P(Y_{new}|y, X|x_{new}, \sigma^2) = Z\, P(Y_{new}|w, X, x_{new}, \sigma^2) P(w|X)\, dw \qquad (12)$$

Predictive statistics analyses the probability of a value $y_0$ set $x_0$ for a specific $w$, by using likelihood. Lastly, add together all the potential values of $w$.

   *b) Neural Network Regression*

Because neural network regression is a supervised learning technique, it necessitates a labeled dataset with a label column [35]. Because a regression model predicts a numerical value, the label column must also be numerical.

You can run the model using Train Model or Tune Hyper-parameters with the model and the tagged dataset as input. After then, the trained model can be utilized to predict values for extra input samples. The following Eq. (13) is used in neural network regression:

$$\sum_{i=0}^{M} w_i x_i = w \cdot x \qquad (13)$$

A perceptron is used in Neural Networks to calculate the linear combinations of inputs given a vector of real-valued inputs. If the output exceeds a certain threshold, it returns 1, else -1. The weights must be computed using the perceptron training criteria (Eq. 14 and 15):

$$w_i \leftarrow w_i + \Delta w_i \qquad (14)$$

$$\Delta w_i = \eta(t - 0)x_i \tag{15}$$

Where $\eta$ denotes to the learning rate of neural network.

    *c) Decision Forest Regression*

Non-parametric decision tree algorithm proceeds through a binary tree structure until they approach a leaf node, conducting a series of simple evaluations on each instance (choice) [36].

This regression model is made up of a collection of decision trees. Each tree in a regression decision forest predicts a Gaussian distribution. The ensemble of trees in the model is aggregating to find a Gaussian distribution that is closest to the mutual distribution for all trees in the model. The Gini value is utilized in Random Forest to determine the tree's final class. The coefficient of the Gini is derived using Eq. 16:

$$Gini(T) = 1 - \sum_{i}^{n}(p_j) \tag{16}$$

Where T dataset is divided into $T_1$, $T_2$ subsets with $N_1$, $N_2$ dimensions, then Gini split value is estimated based on the following Eq. 17:

$$Gini_{split(T)} = \frac{N_1}{N}Gini(T_1) + \frac{N_2}{N}Gini(T_2) \tag{17}$$

    *d) Boosted Decision Tree Regression*

In Azure ML, boosted decision tree is one of the most common predictive algorithms that can be used for both classification and regression [37]. The weighted sum of functions is used in this approach to optimize a differentiable loss function. The $F(x)$ is estimated based on Eq. (18):

$$F_0(x) = argmin \sum_{i}^{n} L(y_i . \gamma) \tag{18}$$

## 5. EXPERIMENTAL RESULTS

The focus of this experiment is on the 'prediction of the received signals in MC system (Map values)' module. However, it is illustrated in a typical Azure regression experiment process, i.e. input data, initialize model, train model, score model, assess model, for a better understanding of the module's function and potential reuse. Figure 4 depicts the experiment workflow.

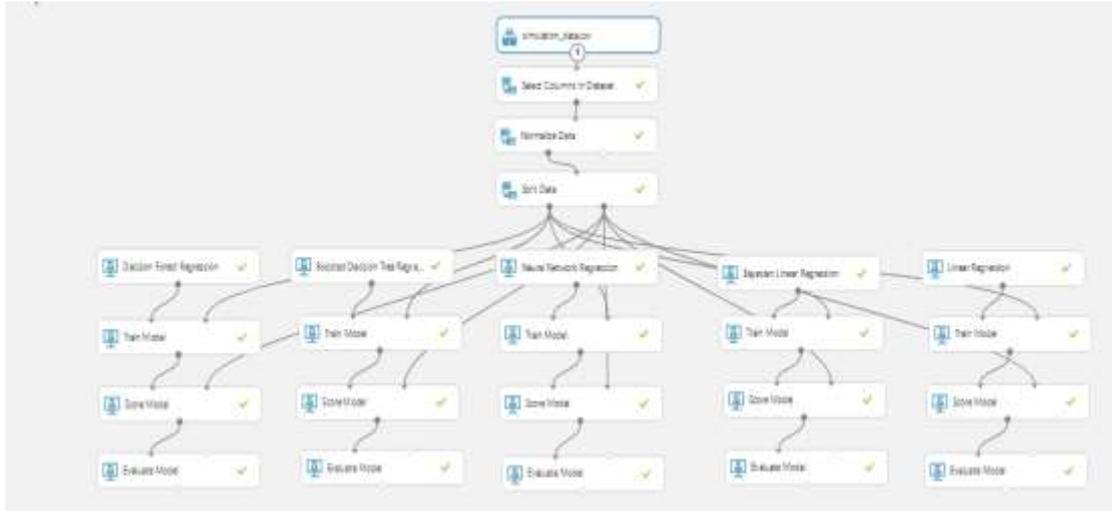

Fig.4 the experiment workflow process.

Datasets are obtained from the simulation study and framework that have been done previously in [32]. There is a file called simulation.csv. In simulation.csv, there are numerous features such as the receiver radius, transmitter radius, the distance between the receiver and the transmitter, and the diffusion coefficient while the output is mAP. Firstly, Create a new experiment as the initial step. Then, the simulation.csv creates a dataset. Feature selection and feature extraction are the next steps. Then, apply data split to the dataset into training and testing dataset. After that, we apply the chosen regression model. Finally, we evaluate the trained regression model's performance. Figure 5 depicts a block diagram of the experiment process.

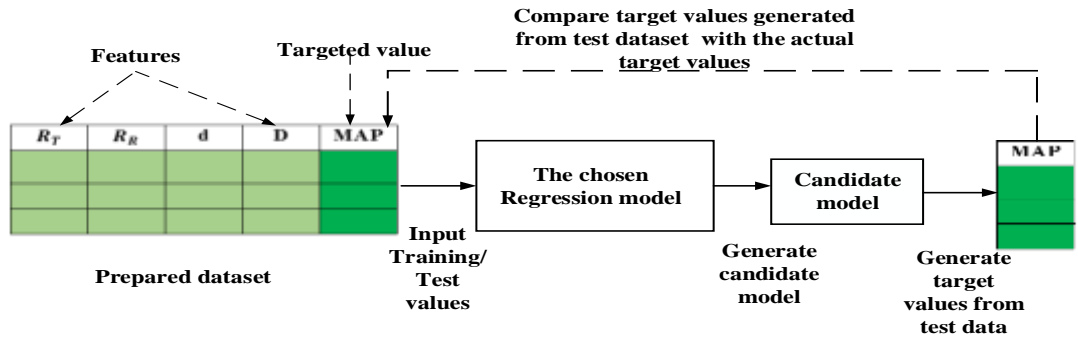

Fig.5 Block diagram of the experiment process.

## 6. Performance metrics

Five commonly used performance metrics are selected for the experiment. The Mathematical definitions of the performance metrics are listed in Table 3.

Table 3 mathematical formulation of the performance metrics used in our work

| Metric name | Metric abbreviation | Metric formula |
|---|---|---|
| Mean Absolute Error | MAE | $MAE = \dfrac{1}{N}\sum_{j=1}^{N} e_j$ |
| Root Mean Squared Error | RMSE | $RMSE = \sqrt{\dfrac{\sum_{j=1}^{N} e_j^2}{N}}$ |
| Relative Absolute Error | RAE | $RAE = \dfrac{\sum_{j=1}^{N}|e_j|}{\sum_{j=1}^{N}|a_{j-\bar{a}}|}$ |
| Relative Squared Error | RSE | $RSE = \sqrt{\dfrac{\sum_{j=1}^{N} e_j^2}{\sum_{j=1}^{N}|a_{j-\bar{a}}|}}$ |
| Coefficient of Determination | CoD | $CoD = 1 - \dfrac{(\sum_{j=1}^{n} p_j - a_j)^2}{(\sum_{j=1}^{n} a_j - \bar{a})^2}$ |

where $a_j$ shows the actual values, $\bar{a}$ is the mean of the actual value, $p_j$ indicates to the predicted values, N is the size of the dataset and $e_j = a_j - p_j$ - $error$ .

## 7. Results

Test results are evaluated by a number of evaluation metrics, for example; MAE, RMSE, RAE, RSE and CoD. Boosted Decision Tree was the best approach based on RMSE, MAE and CoD values. Table 4 shows that the five methods provide good performance.

Table 4 Comparison results

| Metric | Bayesian linear regression | Neural network | Decision forest regression | Boosted decision tree |
|---|---|---|---|---|
| MAE | 0.46368 | 0.089317 | 0.065424 | **0.056942** |
| RMSE | 0.706371 | 0.139959 | 0.139331 | **0.097559** |
| RAE | 0.641535 | 0.123577 | 0.090518 | **0.078783** |
| RSE | 0.413437 | 0.016231 | 0.016086 | **0.007886** |
| CoD | 0.586563 | 0.983769 | 0.983914 | **0.992114** |
| Negative Log-Likelihood | 1587.426653 | - | -770.504254 | - |

The coefficient of determination (CoD) is a number between 0 and 1 that measures the model's predictive power. It's a common method of determining how well a model matches the data. It can be thought of as the percentage of variation that the model can explain. A larger proportion suggests a better fit, with 1 indicating a perfect match. To express our accuracy, we employed the CoD factor. Furthermore, with the Boosted Decision Tree Regression approach, the maximum prediction success in estimates employed is 0.99, as shown in Figure 6.

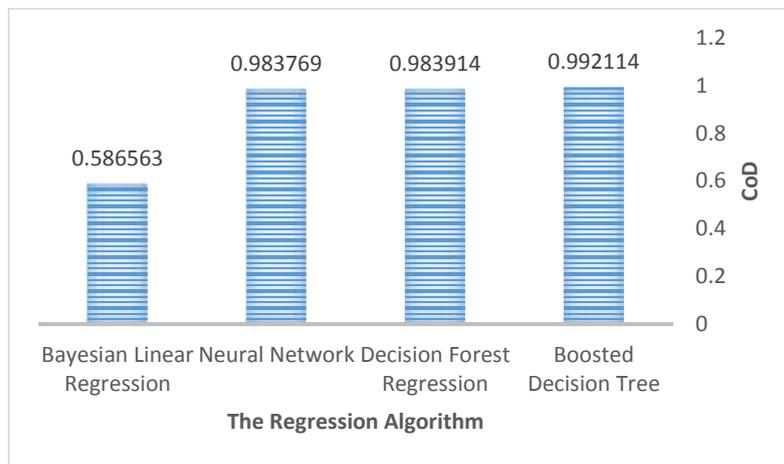

Fig.6 experiment evaluation based CoD.

In the following Figure 7, we show the error histogram of our evaluation in respect to CoD.

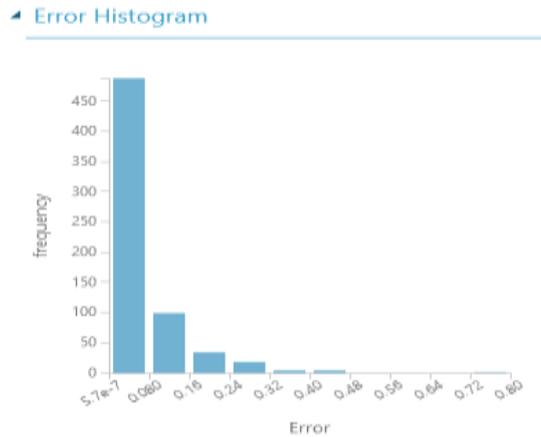

Fig.7 CoD error histogram.

In our previous work [32], different regression approaches (e.g., support vector regression, linear regression, and ridge regression) have been applied for modeling MC received signal until time $t$ using python.

In this work, we notice that regression models (e.g., Boosted Decision Tree Regression) perform significantly better than the algorithms used in [32], and that high performance does not necessitate extensive optimization, as shown in Figure 8. Because we built our NN model with a standard topology, the model's performance may suffer as a result. This can be well thought-out as one of the intimidations to the authority of this work. Furthermore, because our experiments were conducted on a specific dataset, the results may differ on other types of datasets. For modelling MC received signals, we recommend that specialists use the Boosted Decision Tree Regression algorithm.

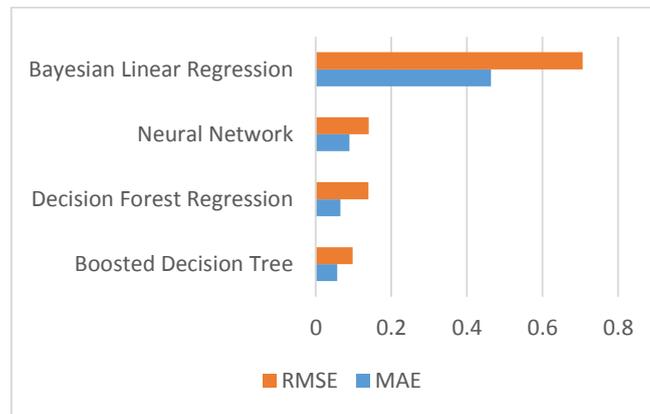

Fig.8 performance evaluation based on RMSE and MAE using Azure ML.

## 8. A comparison between Azure ML and Python algorithms

There has never been a comparison of the performance of Azure ML services vs on premise ML algorithms provided by code-based tools (such as Python). The aim of this work is to compare the performance of two well-known software, Azure ML (visual cloud solution) and Python including ML libraries and tools (**scikit-learn and tensor flow**) (on-basis code solution), on detecting the incoming signal in the MC framework case. The RMSE is the evaluation measures utilized in this work to deliver a primary measureable evaluation of the performances of Azure ML and comparable algorithms in Python.

In the first case, the comparison is based on algorithms that were similar, equally Azure and Python provide the same algorithm. The selected algorithm is Linear Regression(LR). In Table 5 is presented that both Azure ML and Python deliver algorithm with good accuracy.

Table 5 LR based Azure and Python comparison results

| Tool Name | RMSE |
|---|---|
| Azure ML | 0.02066 |
| PYTHON | 0.0291 |

In the following Table 6, we compare the RMSE values of LR for different parameters setting using Azure ML. For solution method, we applied ordinary least square and online gradient descent.

- Ordinary least square is a linear regression approach that is widely utilized. The loss function estimates error as the sum of the squares of the distance between the real value and the anticipated line, then minimizes the squared error to fit the model. This approach assumes a strong linear relationship between the inputs and the dependent variable.
- Gradient descent is a technique for minimizing error at each phase of the model training procedure. Gradient descent has many variants, and its optimization for different learning issues has received a lot of attention. If you select this choice for the Solution method, you can control the step size, learning rate, and other parameters. This choice also allows you to use an inclusive variable path.

Table 6 LR comparison results for different parameter setting using Azure ML

| Solution method | L2 regularization weigh | Learning rate | RMSE |
|---|---|---|---|
| Ordinary least square | 0.001 | - | 0.02066 |
| Ordinary least square | 0.1 | - | 0.026625 |
| Ordinary least square | 0.0001 | - | 0.026625 |
| Online gradient descent | 0.001 | 0.1 | 0.041532 |
| Online gradient descent | 0.01 | 0.1 | 0.046022 |

In the second case, because there is no relationship between the algorithms in the two tools, it was chosen to show which one gives the best results in both cases. In terms of Azure's use of CoD, they are as follows:
_ Boosted Decision Tree Regression: 0.992114
_ Decision Forest Regression: 0. 983914

while in PYTHON using MSE metric the best result given by:
_ Ridge regression: 0.00002
_ deep neural network:0.0028
Both programming tools guarantee good accuracy, though the Azure ML tool is the most accurate and has a faster process.

   **9. Efficiency evaluation between Azure ML and Python tools**

   The research presented in this paper creates a new generation of comparing ML tools rather than only algorithms/techniques. To the best of our knowledge, there has never been a comparison of tool performance in the field of MC. Regular tool testing can provide useful information to the end-user community, helping them to make more informed platform decisions. In Table 7 the comparison between Azure ML and Python tool are presented.

Table 7 Azure ML and Python tool comparison.

|  | Azure ML | Python |
|---|---|---|
| Scope | Domain-specific(ML application) | General purpose (application development) |
| Pricing | Free<br>Monthly payment | Free |
| Language type | Visual programming language(VPL) | Text based language |
| Extension | integrating with text based language like python and R language | Extension using libraries written in Python, C and c++ |
| Ease of use | Use drag and drop instead of need to write code | Need to write script following specific syntax |
| Learning curve | Suits new beginner and experienced programmer | Experienced programmer |
| Speed | Speed: Azure ML is a cloud based tool, so processing is not made with your computer, making the reliability and speed top notch | Processing is depend on the used computer features. |
| features | Classification<br>Regressions<br>Clustering<br>Drag-and-drop function<br>Trained models<br>Web serving publishing<br>Modules<br>Datasets<br>Projects<br>APIs<br>Experiments | Pre-processing<br>Regression<br>Classification<br>Clustering<br>Model selection<br>Dimensionality reduction |

Due to the use of ready-made analysis units provided by Azure ML, the complexity of project development (detection and modeling of the MC received signal) has been significantly reduced when compared to direct Python programming. Also, as a result of the performed detection using Azure ML services, the computation time was reduced compared to using code-based programming tool as Python when solving the same task.

## Conclusions

The application of the Azure ML tool in the detection of the MC received signal is presented for the first time. The performance of ML supplied by two tools, Azure (visual tool) and Python (textual-code base), on an MC system as a research case, is compared. Python is a strong language that allows you to perform complicated tasks and data manipulations, while Azure ML allows you to create ML models quickly and easily with high accuracy. Because of its cloud nature, Azure ML tends to improve its speed over Python-based over time. Among the two solutions we've looked at, Azure ML is the one that may be recommended for researchers who are new to the ML field as opposed to those who are more experienced. The tool is simply incredibly robust, with numerous built-in capabilities and extra functionality that would otherwise need the use of third-party libraries. Python is regarded as ideal for skilled users, but Azure ML requires no programming or coding knowledge.